\documentclass[runningheads]{llncs}
\usepackage{times}
\usepackage{url}
\usepackage{latexsym}
\usepackage{layout}

\usepackage{caption}  

\usepackage{graphicx}  

\begin{document}
\title{Incorporating Chinese Radicals Into Neural Machine Translation: Deeper Than Character Level}
%
\titlerunning{Chinese Radicals in Neural Machine Translation}
%


\author{Lifeng Han\inst{1*} \and
Shaohui Kuang\inst{2*}}

%
\authorrunning{Han and Kuang}

%


\institute{ADAPT Centre, School of Computing, Dublin City University, Dublin, Ireland \\
\email{lifeng.han3@mail.dcu.ie}  \and
NLP Lab, Soochow University, Suzhou, P. R. China\\
\email{shaohuikuang@foxmail.com}}

\maketitle              
\begin{abstract}
  In neural machine translation (NMT), researchers face the challenge of un-seen (or out-of-vocabulary OOV) words translation. To solve this, some researchers propose the splitting of western languages such as English and German into sub-words or compounds. In this paper, we try to address this OOV issue and improve the NMT adequacy with a harder language Chinese whose characters are even more sophisticated in composition. We integrate the Chinese radicals into the NMT model with different settings to address the unseen words challenge in Chinese to English translation. On the other hand, this also can be considered as semantic part of the MT system since the Chinese radicals usually carry the essential meaning of the words they are constructed in. Meaningful radicals and new characters can be integrated into the NMT systems with our models. We use an attention-based NMT system as a strong baseline system. The experiments on standard Chinese-to-English NIST translation shared task data 2006 and 2008 show that our designed models outperform the baseline model in a wide range of state-of-the-art evaluation metrics including LEPOR, BEER, and CharacTER, in addition to BLEU and NIST scores, especially on the adequacy-level translation. We also have some interesting findings from the results of our various experiment settings about the performance of words and characters in Chinese NMT, which is different with other languages. For instance, the fully character level NMT may perform well or the state of the art in some other languages as researchers demonstrated recently, however, in the Chinese NMT model, word boundary knowledge is important for the model learning. \footnote{* parallel authors, ranked by alphabet order. In Proceeding of ESSLLI2018, Sofia, Bulgaria}

\end{abstract}

\section{Introduction}











Machine Translation (MT) has a long history dating from 1950s \cite{Weaver1955} as one topic of artificial intelligence (AI) or intelligent machines. It began with rule-based MT (RBMT) systems that apply human defined syntactic and semantic rules of source and target languages to the machine, to example based MT (EBMT),  statistical MT (SMT),  Hybrid MT (e.g. the combination of RBMT and SMT) and then recent years' Neural MT (NMT) models \cite{Nirenburg1989RBMT,carl2003recent,koehn2009statistical,DBLP:journals/corr/BahdanauCB14}. 

 NMT models treat MT task as encoder-decoder work-flow which is much different from the conventional SMT structure \cite{cho2014learning}. The encoder applies in the source language side learning the sentences into vector representations, while the decoder applies in the target language side generating the words from the target side vectors. Recurrent Neural Networks (RNN) models are usually used for both  encoder and decoder, though there are some researchers employing convolutions neural networks (CNN) like \cite{DBLP:journals/corr/ChoMBB14,kalchbrenner13emnlp}. The hidden layers in the neural nets are designed to learn and transfer the information \cite{neubig2017neural}. 

There were some drawbacks in the NMT models e.g. lack of alignment information between source and target side, and less transparency, etc. To address these, attention mechanism was introduced to the decoder first by \cite{DBLP:journals/corr/BahdanauCB14} to pay interests to part information of the source sentence selectively, instead of the whole sentence always, when the model is doing translation. This idea is similar like alignment functions in SMT and what the human translators usually perform when they undertake the translation task. Earlier, attention mechanisms were applied in neural nets for image processing tasks \cite{NIPS2010_4089,Denil2011NIPSattention}. Recently, Attention based models have appeared in most of the NMT projects, such as the the investigation of global attention-based architectures \cite{DBLP:journals/corr/LuongPM15} and target information \cite{peter2017generating} for pure text NMT, and the exploration of Multi-modal NMT \cite{Huang2016AttentionMNMT}. To generalize the attention mechanism in the source language side, coverage model is introduced to balance the weights of different parts of the sentences into NMT by \cite{DBLP:journals/corr/TuLLLL16,Mi2016CoverageEm}.



Another drawback of NMT is that the NMT systems usually produce better fluent output, however, the adequacy is lower sometimes compared with the conventional SMT, e.g. some meaning from the source sentences will be lost in the translation side when the sentence is long \cite{DBLP:journals/corr/TuLLLL16a,DBLP:journals/corr/TuLLLL16,koehn2017six,neubig2017neural,DBLP:journals/corr/ChoMBB14}. One kind of reason of this phenomenon could be due to the unseen words problem, except for the un-clear learning procedure of the neural nets. With this assumption, we try to address the unseen words or out-of-vocabulary (OOV) words issue and improve the adequacy level by exploring the Chinese radicals into NMT.

For Chinese radical knowledge, let's see two examples about their  construction in the corresponding characters. This Figure 1 shows three Chinese characters (forest, tree, bridge) which contain the same part of radical (wood) and this radical can be a character independently in usage. In the history, Chinese bridge was built by wood usually, so apparently, these three characters carry the similar meaning that they all contain something related with woods.

\begin{figure}[!t]
\centering
\includegraphics*[height=1.5in,width=2in]{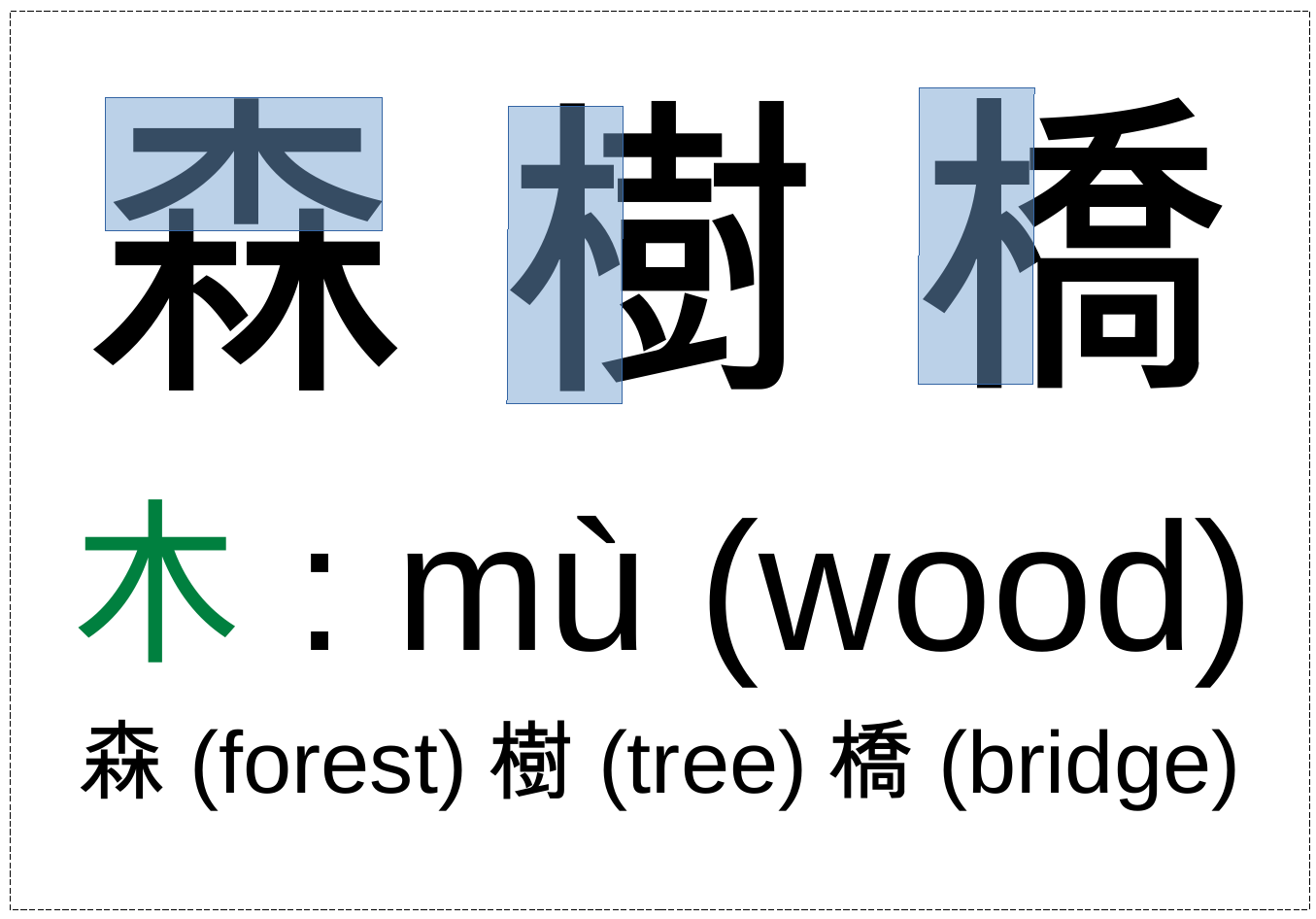}
\caption{Radical as independent character.}
\label{fig:3}
\end{figure}

\begin{figure}[!t]
\centering
\includegraphics*[height=1.5in,width=2in]{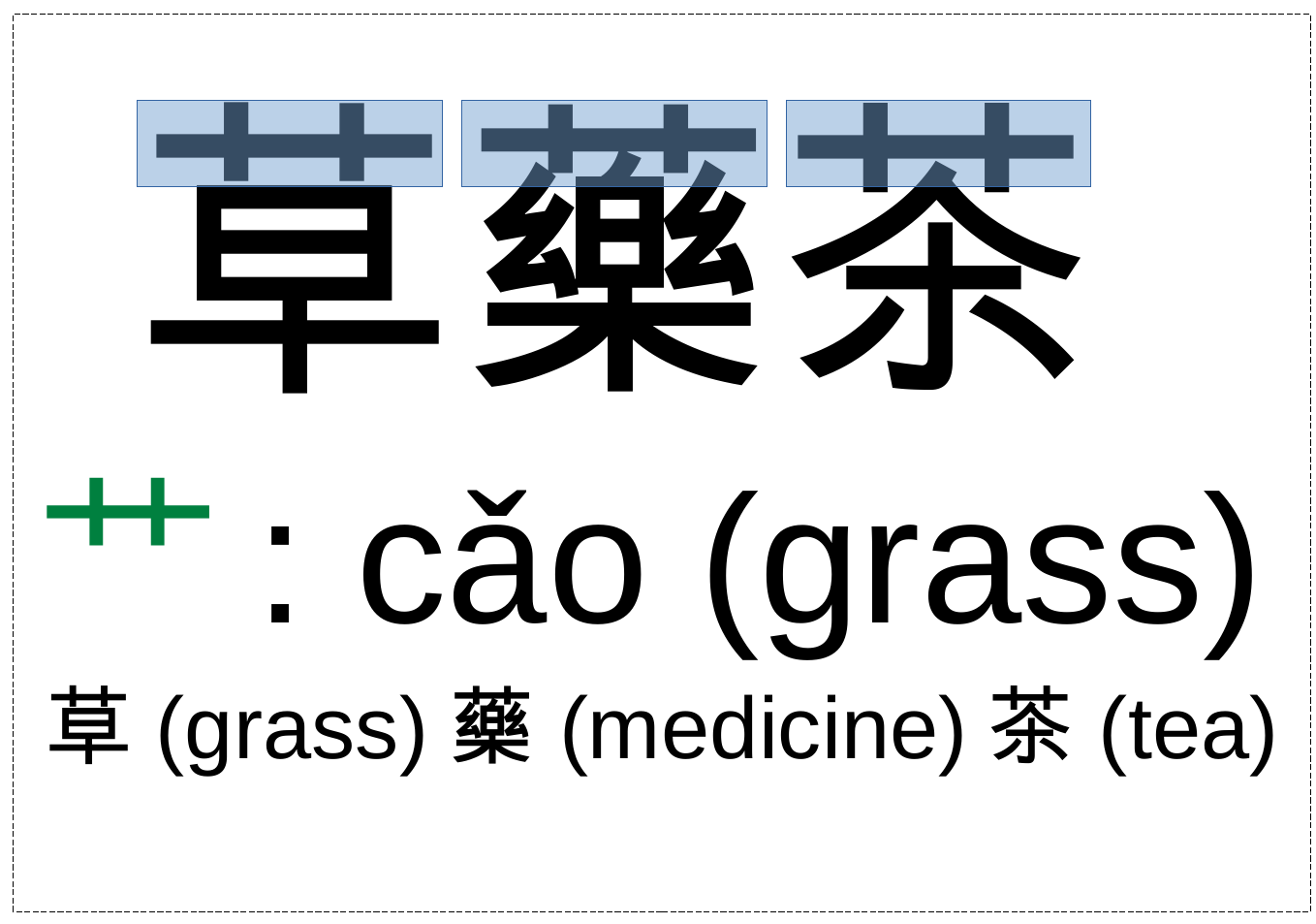}
\caption{Radical as non-independent character.}
\label{fig:3}
\end{figure}

The Figure 2 shows three Chinese characters (grass, medicine, tea) which contain the same part of radical (grass) however this radical can not be a character independently in usage. This radical means grass in the original development of Chinese language. In the history, Chinese medicine was usually developed from some nature things like the grass, and Chinese tea was usually from the leafs that are related with grass. To the best knowledge of the authors at the submission stage, there is no published work about radical level NMT for Chinese language yet. 


\section{Related Work}


MT models have been developed by utilizing smaller units, i.e. phrase-level to word-level, sub-word level and character-level \cite{SubwordNMT15Sennrich,chung2016character}. However, for Chinese language, sub-character level or radical level is also a quite interesting topic since the Chinese radicals carry somehow essential meanings of the Chinese characters that they are constructed in. Some of the radicals splitted from the characters can be independent new characters, meanwhile, there are some other radicals that can not be independent as characters though they also have meanings. It would be very interesting to see how these radicals or the combination of them and traditional words/characters perform in the NMT systems.




There are some published works about the investigation of Chinese radicals embedding for other tasks of NLP, such as  \cite{Radical15ShiNLP,Liu17CompositionalityVisual} explored the radical usage for word segmentation and text categorization. 


Some MT researchers explored the word composition knowledge into the systems, especially on the western languages. For instance, \cite{matthews2016synthesizing} developed a Machine Translation model on English-German and English-Finnish with the consideration of synthesizing compound words. This kind of knowledge is similar like the splitting Chinese character into new characters. 

\section{Model Design}


\subsection{Attention-based NMT}

Typically, as mentioned before, neural machine translation (NMT) builds on an encoder-decoder framework \cite{DBLP:journals/corr/BahdanauCB14,sutskever2014sequence} based on recurrent neural networks (RNN). In this paper, we take the NMT architecture proposed by \cite{DBLP:journals/corr/BahdanauCB14}.  In NMT system, the encoder apples a bidirectional RNN to encode a source sentence $x=(x_1, x_2, ..., x_{T_x})$ and repeatedly generates the hidden vectors $h = (h_1, h_2, ..., h_{T_x})$ over the source sentence, where $T_x$ is the length of source sentence.  Formally, $h_j = [\overrightarrow{h_j};\overleftarrow{h_j}]$ is the concatenation of forward RNN hidden state $\overrightarrow{h_j}$ and backward RNN hidden state $\overleftarrow{h_j}$, and $\overrightarrow{h_j}$ can be computed as follows:

\begin{equation}
    \overrightarrow{h_j} = f(\overrightarrow{h_{j-1}}, x_j)
\end{equation}
where function f is defined as a Gated Recurrent Unit (GRU) \cite{chung2014empirical}.

The decoder is also an RNN that predicts the next word $y_t$ given the context vector $c_t$, the hidden state of the decoder $s_t$ and the previous predicted word $y_{t-1}$, which is computed by:

\begin{equation}
    p(y_t|y_{<t},x) = softmax(g(s_t, y_{t-1}, c_t))
\end{equation}
where $g$ is a non-linear function. and $s_t$ is the state of decoder RNN at time step $t$, which is calculated by:

\begin{equation}
    s_t = f(s_{t-1}, y_{t-1}, c_t)
\end{equation}
where $c_t$ is the context represent vector of source sentence.

Usually $c_t$ can be obtained by attention model and calculated as follows:

\begin{equation}
    c_t = \sum^{T_x}_{j=1} \alpha_{tj}h_j
\end{equation}


\begin{equation}
    \alpha_{tj} = \frac{exp(e_{tj})}{\sum^{T_x}_{k=1}{e_{tk}}}
\end{equation}

\begin{equation}
    e_{tj} = v^T_atanh(s_{t-1},h_j)
\end{equation}

We also follow the implementation of attention-based NMT of dl4mt tutorial \footnote{github.com/nyu-dl/dl4mt-tutorial/tree/master/ session2}, which enhances the attention model by feeding the previous word $y_{t-1}$ to it, therefore the $e_{tj}$ is calculated by:

\begin{equation}
   e_{tj} = v^T_atanh(\widetilde s_{t-1},h_j) 
\end{equation}
where $\widetilde s_{t-1} = f(s_{t-1},y_{t-1})$, and $f$ is a GRU function. The hidden state of the decoder is updated as following:

\begin{equation}
    s_t = f(\widetilde s_{t-1}, c_t) 
\end{equation}

In this paper, we use the attention-based NMT with the changes from dl4mt tutorial \footnote{github.com/nyu-dl/dl4mt-tutorial} as our baseline and call it RNNSearch*\footnote{To distinguish it from RNNSearch as in the paper \cite{DBLP:journals/corr/BahdanauCB14}}.

\subsection{Our model}

Traditional NMT model usually uses the word-level or character-level information as the inputs of encoder, which ignores some knowledge of the source sentence, especially for Chinese language. Chinese words are usually composed of multiple characters, and characters can be further splitted into radicals. The Chinese character construction is very complected, varying from upper-lower structure, left-right structure, to inside-outside structure and the combination of them. In this paper, we use the radical, character and word as multiple inputs of NMT and expect NMT model can learn more useful features based on the different levels of input integration.

\begin{figure}[!t]
\centering
\includegraphics*[height=2.5in,width=1.7in]{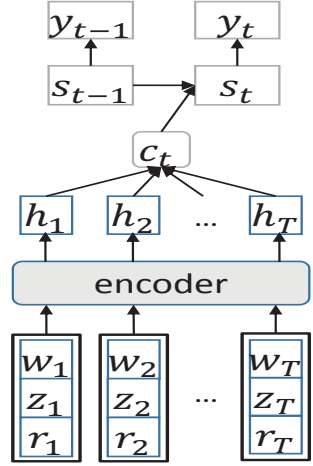}
\caption{Architecture of NMT with multi-embedding.}
\label{fig:3}
\end{figure}

Figure 3 illustrates our proposed model. The input embedding $x_j$ consists of three parts: word embedding $w_j$, character embedding $z_j$ \footnote{We use the character `z' to represent character, instead of `c', because we already used `c' as representation of context vector.} and radical embedding $r_j$, as follows:

\begin{equation}
    x_j = [w_j;z_j;r_j]
\end{equation}
where `;' is concatenate operation.

For the word $w_j$, it can be split into characters $z_j = (z_{j1}, z_{j2}, ..., z_{jm})$ and further split into radicals $r_j = (r_{j1}, r_{j2},..., r_{jn})$. In our model, we use simple additions operation to get the character representation and radical representation of the word, i.e. $z_j$ and $r_j$ can be computed as follows:

\begin{equation}
    z_j = \sum_{k=1}^{m} z_{jk}
\end{equation}
\begin{equation}
     r_j = \sum_{k=1}^n r_{jk}
\end{equation}

Each word can be decomposed into different numbers of character and radical, and, by addition operations, we can generate a fixed length representation. In principle our model can handle different levels of input from their combinations. For Chinese character decomposition, e.g. the radicals generation, we use the HanziJS open source toolkit \footnote{github.com/nieldlr/Hanzi}. On the usage of target vocabulary \cite{jean2014using}, we choose 30,000 as the volume size.

\section{Experiments}

\subsection{Experiments Setting}

We used 1.25 million parallel Chinese-English sentences for training, which contain 80.9 millions Chinese words and 86.4 millions English words. The data is mainly from Linguistic Data Consortium (LDC) \footnote{www.ldc.upenn.edu} parallel corpora, such as LDC2002E18, LDC2003E07, LDC2003E14, LDC2004T07, LDC2004T08, and LDC2005T06. We tune the models with NIST06 as development data using BLEU metric \cite{Papineni02bleu:a}, and use NIST08 Chinese-English parallel corpus as testing data with four references.


For the baseline model RNNSearch*, in order to effectively train the model, we limit the maximum sentence length on both source and target side to 50. We also limit both the source and target vocabularies to the most frequent 30k words and replace rare words with a special token ``UNK'' in Chinese and English. The vocabularies cover approximately 97.7\% and 99.3\% of the two corpora, respectively.  Both the encoder and decoder of RNNsearch* have 1000 hidden units. The encoder of RNNsearch consists of a forward (1000 hidden unit) and backward bidirectional RNN. The word embedding dimension is set as 620. We incorporate dropout \cite{hinton2012improving} strategy on the output layer. We used the stochastic descent algorithm with mini-batch and Adadelta \cite{zeiler2012adadelta} to train the model. The parameters $\rho$ and $\epsilon$ of Adadelta are set to 0.95 and $10^{-6}$. Once the RNNsearch* model is trained, we adopt a beam search to find possible translations with high probabilities. We set the beam width of RNNsearch* to 10. The model parameters are selected according to the maximum BLEU score points on the development set.

For our proposed model, all the experimental settings are the same as RNNSearch*, except for the word-embedding dimension and the size of the vocabularies. In our model, we set the word, character and radical to have the same dimension, all 620. The vocabulary sizes of word, character and radical are set to 30k, 2.5k and 1k respectively. 

To integrate the character radicals into NMT system, we designed several different settings as demonstrated in the table. Both the baseline and our settings used the attention-based NMT structure.

\begin{center}
\captionof{table}{Model Settings}
\begin{tabular}{ l|c |  c } \hline 
  Settings & Description & abbreviation \\ \hline 
  Baseline & Words & W \\ 
  Setting1 & Word+Character+Radical & W+C+R \\
  Setting2 & Word+Character & W+C \\
  Setting3 & Word+Radical & W+R \\
  Setting4 & Character+Radical & C+R \\ \hline
  
\end{tabular}
\end{center}

\subsection{Evaluations}

Firstly, there are many works reflecting the insufficiency of BLEU metric, such as higher or lower BLEU scores do not necessarily reflect the model quality improvements or decreasing; BLEU scores are not interpretive by many translation professionals; and BLEU did not correlate better than later developed metrics in some language pairs \cite{callison2006re,callison2007meta,lavie2013automated}.

In the light of such analytic works, we try to validate our work in a deeper and broader evaluation setting from more aspects. We use a wide range of state of the art MT evaluation metrics, which are developed in recent years, to do a more comprehensive evaluation, including hLEPOR \cite{han2013language,han2014lepor}, CharacTER \cite{wang2016character}, BEER \cite{stanojevic-simaan:2014:W14-33}, in addition to BLEU and NIST \cite{Papineni02bleu:a}.  

The model hLEPOR is a tunable translation  evaluation metric yielding higher correlation with human judgments by adding n-gram position difference penalty factor into the traditional F-measures. CharacTER is a character level editing distance rate metric. BEER uses permutation trees and character n-grams integrating many features such as paraphrase and syntax. They have shown top performances in recent years' WMT\footnote{www.statmt.org/wmt17/metrics-task.html} shared tasks  \cite{machavcek-bojar:2013:WMT,machacek-bojar:2014:W14-33,grahametal:15,bojar-EtAl:2016:WMT2}.

Both CharacTER and BEER metrics achieved the parallel top performance in correlation scores with human judgment on Chinese-to-English MT evaluation in WMT-17 shared tasks \cite{bojar-graham-kamran:2017:WMT} .
While LEPOR metric series are evaluated by MT researchers as one of the most distinguished metric families that are not apparently outperformed by others, which is stated in the metrics comparison work in \cite{grahametal:15} on standard WMT data.

\subsubsection{Evaluation on Development Set}

On the development set NIST06, we got the following evaluation scores. The cumulative N-gram scoring of BLEU and NIST metric, with bold case as the highlight of the winner in each n-gram column situation, is shown in the table respectively. Researchers usually report their 4-gram BLEU while 5-gram NIST metric scores, so we also follow this tradition here:

\begin{center}
\captionof{table}{BLEU Scores on NIST06 Development Data}
\begin{tabular}{ l|c |c |c | c } \hline 
   & 1-gram & 2-gram & 3-gram & 4-gram   \\ \hline 
  Baseline & .7211 & .5663 & .4480 & .3556  \\
  W+C+R & \textbf{.7420} & \textbf{.5783} & \textbf{.4534} & \textbf{.3562} \\
  W+C & .7362 & .5762 & .4524 & .3555  \\
  W+R & .7346 & .5730 & .4491 &  .3529  \\
  C+R & .7089 & .5415 & .4164 & .3219  \\ \hline 
\end{tabular}
\end{center}


\begin{center}
\captionof{table}{NIST Scores on NIST06 Development Data}
\begin{tabular}{ l|c |c |c | c| c } \hline 
   & 1-gram & 2-gram & 3-gram & 4-gram & 5-gram  \\ \hline
  Baseline & 5.8467 & 7.7916 & 8.3381 & 8.4796 & 8.5289  \\
  W+C+R & \textbf{6.0047}  & \textbf{7.9942} & \textbf{8.5473} & \textbf{8.6875} & \textbf{8.7346}  \\
  W+C & 5.9531 & 7.9438 & 8.5127 & 8.6526 & 8.6984  \\
  W+R & 5.9372 & 7.9021 & 8.4573 & 8.5950 & 8.6432  \\
  C+R & 5.6385 & 7.4379 & 7.9401 & 8.0662 & 8.1082  \\ \hline 
\end{tabular}
\end{center}

From the scoring results, we can see that the model setting one, i.e. W+C+R, won the baseline models in all uni-gram to 4-gram BLEU and to 5-gram NIST scores. Furthermore, we can see that, by adding character and/or radical to the words, the model setting two and three also outperformed the baseline models. However, the setting 4 that only used character and radical information in the model lost both BLEU and NIST scores compared with the word-level baseline. This means that, for Chinese NMT, the word segmentation knowledge is important to show some guiding in Chinese translation model learning.

For uni-gram BLEU score, our Model one gets 2.1 higher score than the baseline model which means by combining W+C+R the model can yield higher adequacy level translation, though the fluency score (4-gram) does not have much difference. This is exactly the point that we want to improve about neural models, as complained by many researchers.


The evaluation scores with broader state-of-the-art metrics are shown in the following table. Since CharacTER is an edit distance based metric, the lower score means better translation result.

\begin{center}
\captionof{table}{Broader Metrics Scores on NIST06 Development Data}
\begin{tabular}{ l|c  |c |  c } \hline 
   & \multicolumn{3}{c}{Metrics on Single Reference}  \\ \hline 
  Models  & hLEPOR & BEER & CharacTER \\ \hline
  Baseline & .5890 & .5112 & .9225  \\
  W+C+R  & .5972 & \textbf{.5167} & \textbf{.9169}  \\
  W+C  & \textbf{.5988} & .5164 & .9779  \\
  W+R  & .5942 & .5146 &  .9568 \\
  C+R  & .5779 & .4998 & 1.336  \\ \hline 
\end{tabular}
\end{center}

From the broader evaluation metrics, we can see that our designed models also won the baseline system in all the metrics. Our model setting one, i.e. the W+C+R model, won both BEER and CharacTER scores, while our model two, i.e. the W+C, won the hLEPOR metric score, though the setting four continue to be the worest performance, which is consistent with the BLEU and NIST metrics. Interestingly, we find that the CharacTER score of setting two and three are both worse than the baseline, which means that by adding of character and radical information separately the output translation needs more editing effort; however, if we add both the character and radical information into the model, i.e. the setting one, then the editing effort became less than the baseline. 


\subsubsection{Evaluation on Test Sets}
The evaluation results on the NIST08 Chinese-to-English test date are presented in this section.

Firstly, we show the evaluation scores on BLEU and NIST metrics, with four reference translations and case-insensitive setting. The tables show the cumulative N-gram scores of BLEU and NIST, with bold case as the winner of each n-gram situation in each column.

\begin{center}
\captionof{table}{BLEU Scores on NIST08 Test Data}
\begin{tabular}{ l|c |c |c | c} \hline 
   & 1-gram & 2-gram & 3-gram & 4-gram   \\ \hline 
  Baseline & .6451 & .4732 & .3508 & .2630  \\
  W+C+R & \textbf{.6609} & \textbf{.4839} & \textbf{.3572} & \textbf{.2655} \\
  W+C & .6391 & .4663 & .3412 & .2527 \\
  W+R & .6474 & .4736 & .3503 &  .2607  \\
  C+R & .6378 & .4573 & .3296 & .2410  \\ \hline 
\end{tabular}
\end{center}

\begin{center}
\captionof{table}{NIST Scores on NIST08 Test Data}
\begin{tabular}{ l|c |c |c | c| c } \hline 
   & 1-gram & 2-gram & 3-gram & 4-gram & 5-gram  \\ \hline
  Baseline & 5.1288 & 6.6648 & 7.0387 & 7.1149 & 7.1387  \\
  W+C+R & \textbf{5.2858}  & \textbf{6.8689} & \textbf{ 7.2520} & \textbf{7.3308} & \textbf{7.3535}  \\
  W+C & 5.0850 &  6.5977 & 6.9552 & 7.0250 & 7.0467  \\
  W+R & 5.1122 & 6.6509 & 7.0289 & 7.1062 & 7.1291  \\
  C+R & 5.0140 & 6.4731 & 6.8187 & 6.8873 & 6.9063  \\ \hline 
\end{tabular}
\end{center}

The results show that our model setting one won both BLEU and NIST scores on each n-gram evaluation scheme, while model setting three, i.e. the W+R model, won the uni-gram and bi-gram BLEU scores, and got very closed score with the baseline model in NIST metric. Furthermore, the model setting four, i.e. the C+R one, continue showing the worst ranking, which may verify that word segmentation information and word boundaries are indeed helpful to Chinese translation models, so we can not omit such part.

What worth to mention is that the detailed evaluation scores from BLEU reflect our Model one yields higher BLEU score (1.58) on uni-gram, similar with the results on development data, while a little bit higher performance on 4-gram (0.25). These mean that in the fluency level our translation is similar with the state-of-the-art baseline, however, our model yields much better adequacy level translation in NMT since uni-gram BLEU reflects the adequacy aspect instead of fluency. This verifies the value of our model in the original problem we want to address.

The evaluation results on recent years' advanced metrics are shown below. The scores are also evaluated on the four references scheme. We calculate the average score of each metric from 4 references as the final evaluation score. Bold case means the winner as usual.

\begin{center}
\captionof{table}{Broader Metrics Scores on NIST08 Test Data}
\begin{tabular}{ l|c |c  |  c } \hline 
   & \multicolumn{3}{c}{Metrics Evaluated on 4-references}  \\ \hline 
  Models  & hLEPOR & BEER & CharacTER \\ \hline
  Baseline  & .5519 & .4748 & \textbf{0.9846} \\
  W+C+R & \textbf{.5530} & \textbf{.4778} & 1.3514  \\
  W+C &  .5444 & .4712 & 1.1416  \\
  W+R &  .5458 & .4717 &  0.9882 \\
  C+R &  .5353 & .4634 & 1.1888  \\ \hline 
\end{tabular}
\end{center}

From the broader evaluations, we can see that our model setting one won both the LEPOR and BEER metrics. Though the baseline model won the CharacTER metric, the margin between the two scores from baseline (.9846) and our model three, i.e. W+R, (.9882) is quite small around 0.0036. Continuously, the setting four with C+R performed the worst though and verified our previous findings.

\section{Conclusion and Future Work}

We presented the different performances of the multiple model settings by integrating Chinese character and radicals into state-of-the-art attention-based neural machine translation systems, which can be helpful information for other researchers to look inside and gain general clues about how the radical works.

Our model shows the full character+radical is not enough or suitable for Chinese language translation, which is different with the work on western languages such as \cite{chung2016character}. Our model results showed that the word segmentation and word boundary are helpful knowledge for Chinese translation systems.

Even though our model settings won both the traditional BLEU and NIST metrics, the recent years developed advanced metrics indeed showed some differences and interesting phenomena, especially the character level translation error rate metric CharacTER. This can encourage MT researchers to use the state-of-the-art metrics to find useful insight of their models.

Although the combination of words, characters and radicals mostly yielded the best scores, the broad evaluations also showed that the model setting W+R, i.e. using both words and radicals information, is generally better than the model setting W+C, i.e. words plus characters without radical, which verified the value of our work by exploring radicals into Chinese NMT. Our Model one yielded much better adequacy level translation output (by uni-gram BLEU score) compared with the baseline system, which also showed that this work is important in exploring how to improve adequacy aspect of neural models. 

In the future work, we will continue to optimize our models and use more testing data to verify the performances. In this work, we aimed at exploring the effectiveness of Chinese radicals, so we did not use BPE for English side splitting, however, to promote the state-of-the-art Chinese-English translation, in our future extension, we will apply the splitting on both Chinese and English sides. We will also investigate the usage of Chinese radicals into MT evaluation area, since they carry the language meanings.

\section{Acknowledgement}
The author Han thanks Ahmed Abdelkader for the kind help, and Niel de la Rouviere for the HanziJS toolkit. This work was supported by Soochow University of China and ADAPT Centre of Ireland. The ADAPT Centre for Digital Content Technology is funded under the SFI Research Centres Programme (Grant 13/RC/2106) and is co-funded under the European Regional Development Fund.

%
%
%
%
 \bibliographystyle{splncs04}
%

\bibliography{mtsummit2015}

\end{document}